\documentclass[conference]{IEEEtran}
\IEEEoverridecommandlockouts
\usepackage{cite}
\usepackage{amsmath,amssymb,amsfonts}
\usepackage{algorithmic}
\usepackage{graphicx}
\usepackage{textcomp}
\usepackage{xcolor}
\usepackage{booktabs}
\def\BibTeX{{\rm B\kern-.05em{\sc i\kern-.025em b}\kern-.08em
    T\kern-.1667em\lower.7ex\hbox{E}\kern-.125emX}}
\begin{document}

\title{Language Independent Sentiment Analysis\\
}



\author{
	\IEEEauthorblockN{Muhammad Haroon Shakeel\IEEEauthorrefmark{1}, Safi Faizullah\IEEEauthorrefmark{2}, Turki Alghamidi\IEEEauthorrefmark{3}, Imdadullah Khan\IEEEauthorrefmark{1}}
	\IEEEauthorblockA{\IEEEauthorrefmark{1}\textit{Department of Computer Science, Lahore University of Management Sciences (LUMS),} \\ Lahore, Pakistan
		\\ \IEEEauthorrefmark{1}\{m.shakeel,imdad.khan\}@lums.edu.pk}
	\IEEEauthorblockA{\IEEEauthorrefmark{2}\IEEEauthorrefmark{3}\textit{Department of Computer Science, Islamic University,} \\ Madina, Saudi Arabia
		\\ \IEEEauthorrefmark{2} safi@iu.edu.sa, \IEEEauthorrefmark{3}dr.turki.iu@gmail.com}

}
	
\maketitle

\begin{abstract}
Social media platforms and online forums generate a rapid and increasing amount of textual data. Businesses, government agencies, and media organizations seek to perform sentiment analysis on this rich text data. The results of these analytics are used for adapting marketing strategies, customizing products, security, and various other decision makings. Sentiment analysis has been extensively studied and various methods have been developed for it with great success. These methods, however, apply to texts written in a specific language. This limits the applicability to a particular demographic and geographic region. In this paper, we propose a general approach for sentiment analysis on data containing texts from multiple languages. This enables all the applications to utilize the results of sentiment analysis in a language oblivious or language-independent fashion. 
\end{abstract}   



\section{Introduction}
In recent years, there has been a tremendous increase in generating textual data. Publicly available texts are produced in the form of posts on social media platforms, questions and answers on forums, comments, reviews of products, news, and policy proposals to name a few.  The information content of this data is very rich and is of great utility to various sectors of society and economy. Inferring the opinion or feelings of the author of the texts (referred to as sentiment of the text) is of great value for the government, manufacturers, retailers, media organizations, academia, and many other entities.

Manufacturers and content producers can modify product designs or customize according to the market feedback inferred from sentiments of the reviews and comments provided by the users. This, in turn, can lead to greater customer satisfaction and hence enhanced revenue. Advertisers and marketing agencies can use this information for adapting effective marketing strategies and optimize their budgets. Governments and policymakers could use it to predict social reactions to a policy proposal. Sports organizations and clubs can benefit from sentiment analysis on fans' comments to measure players' contribution to a game and hence estimate players' worth. 

Given the vast volume, velocity, and varying writing styles of the text, performing manual sentiment analysis, even on a small subset of texts is nearly impossible. Various methods using tools from text analytics, natural language processing and computational linguistics have been developed to automatically judge the sentiment of the text and hence the mood of the source of the text. These tools, however are mainly applicable to the text datasets written in a specific language. Moreover, a wide majority of sentiment analysis models work on the English language. Although English is the most common language used over the internet, there is a sizable body of texts that is written in various other languages.  

A versatile sentiment analysis tool that is language-independent will open a broad avenue of applications that are not limited to a specific geographic region or a  group of people. It will also essentially use the best of all worlds, particularly the wide array of tools developed by the natural language processing and text analytics community for the English language can be extended to other languages. 

In this paper, we address the problem of sentiment analysis on short texts. We evaluate our model on two publicly available datasets comprised of English and German languages. Our model predicts the sentiment of a text while being oblivious to the language of the text. We achieve significant improvement over the current state-of-the-art on these two dataset in terms of language-independent sentiment analysis.

We report the results of experiments on mixed as well as individual languages. Our model is straight-forward and does not incorporate any language-specific tool. Thus it can readily be extended for language-independent sentiment analysis with more than two languages.

The rest of the paper is organized as follows. We provide a brief review of existing methods for sentiment analysis, multilingual sentiment analysis, and language-independent sentiment analysis in Section \ref{relatedwork}. The proposed model is described in Section \ref{sec:proposedApproach}. The description of the datasets along with the results of the experiments are presented in Section \ref{sec:evaluation}. Finally, we conclude the paper in Section \ref{sec:conclusion}.

\section{Related Work}\label{relatedwork}

People tend to share ideas, opinions, and emotions towards a particular movie, product, service, or personality on the internet. Such shares are often posted on social media platforms. These posts are a source of firsthand information, such as tweets from people at a concert, a game, or a conference. Automatic detection of emotions in these posts can give insight to what populous feel about a particular event, personality or product, consequently facilitating managerial decisions. The problem of identification of emotion in a text is known as ``sentiment analysis". 

Detecting sentiment in a longer text, such as movie reviews \cite{Turney_2002_b1}, is easier as compared to shorter text due to the greater available context. In tweets, for instance, people tend to use self-created abbreviations due to tweet-length restriction of $140$ characters. Furthermore, short texts are also influenced by regional languages and create a distinct informal dialect of communication \cite{Rafae_2015_b2}. The informal language, spelling variations of the same word, and self-created abbreviations make the automatic sentiment analysis task very challenging. One way to enhance short text sentiment analysis is to normalize lexical variations of a word to its standard root word \cite{Khan_2012_b3}. However, such an approach requires external resources, which might not be available for all languages. More sophisticated approaches translate the source language into English and utilize readily available resources of English language to perform sentiment analysis~\cite{Zhou_2016_b4}. However, such an approach cannot be generalized across all languages due to non-availability of robust translation tools.
In recent years, deep learning has been extensively used to understand and model natural language. No complex feature engineering is required to model the richness of human language~\cite{Denecke_2008_b5}. For language-independent sentiment analysis, a deep learning network has been proposed in \cite{Medrouk__2017_b6}, which utilizes convolutional neural network (CNN) for feature learning and classification. They proved language independence of the model on English, French, and Greek languages. It is established that using pre-trained embeddings can increase the performance of a deep learning model for tasks involving NLP~\cite{subramani2019deep, shakeel2019multi}. Such embeddings, trained on large scale corpus, are readily available for English language~\cite{pennington-etal-2014-glove,peters-etal-2018-deep}. However, not every language enjoys the luxury of the availability of such resources. Therefore, more recent approaches avoid using any language-specific features such as dictionaries, morphological structures, syntactic preprocessing, or pre-trained embeddings~\cite{Attia_2018_b9}. They show that their approach achieves state of the art on three datasets comprising of English, German, and Arabic languages.

Another issue with pre-trained embedding is that it does not account for sentiment polarity of words and might map words with opposite polarities to vectors closer to each other in Euclidean space. A novel sentiment-specific word embedding is proposed for language-independent sentiment analysis, which shows an improvement over traditional pre-trained embeddings of word2vec\cite{Saroufim_2018_b8}.

The most recent approach addresses the problem of language and domain-independent sentiment detection using the surrounding context of the analyzed document in a character n-gram based model \cite{Kincl_2019_b10}. Their all-in-one classifier achieves comparable performance on multiple domains and languages.

\section{Proposed Approach} \label{sec:proposedApproach}

For given input text, each word is first mapped to a $d$ dimensional vector, known as word embedding. This word embedding is later used to learn features representations. These features are forwarded to a small classifier for final prediction. More formally, given a text $T$ with $t$ words $(T = \;<w_1, w_2, w_3, ... , w_t>)$, where each word is defined by a $d$ dimensional vector initialized randomly from a uniform distribution, our goal is to predict sentiment $y\in \{\text{positive, neutral, negative}\}$ for $T$. We propose a Convolutional Neural Network (CNN) and Long Short-term Memory (LSTM) based architecture to achieve this goal.

\begin{figure}[!bp]
	\centering
	\includegraphics[page=1, scale=0.65, trim={135 135 140 140},clip]{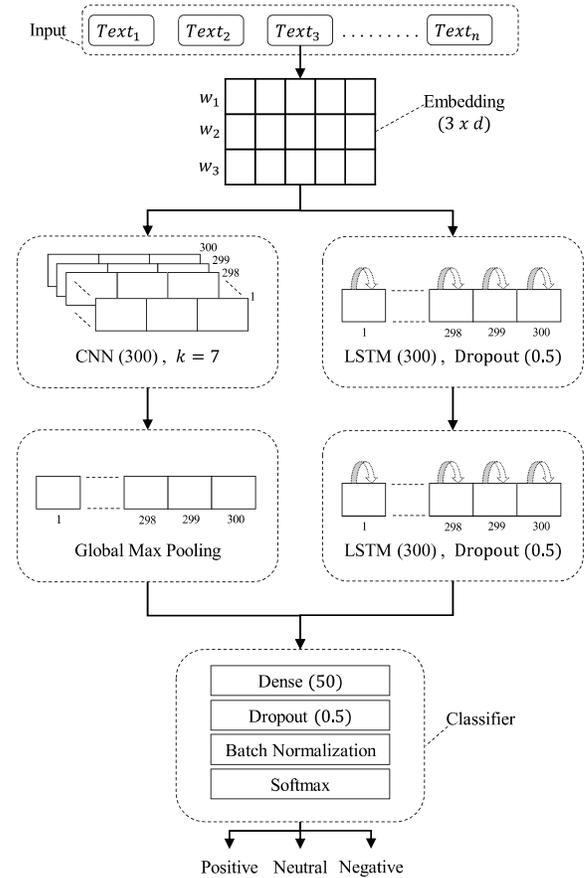}
	\caption{Language independent sentiment analysis model architecture}
	\label{fig:architecture}
\end{figure}

The choice of CNN is made to learn $n$-gram features from the input text as it has been proven that utilizing $n$-gram features are useful for language-independent models~\cite{Medrouk__2017_b6, Attia_2018_b9}. However, these existing architectures ignore the fact that encoding long-term dependencies in text is also crucial for learning robust feature representation~\cite{Zhou_2016_b4}. In the proposed model, we utilize two layers of LSTM to capture long-term dependencies, while to learn $n$-gram features, CNN is used. A global max-pooling is used on the representation learned by CNN, which outputs the maximum value from each feature map learned by a particular filter. The outputs of the second LSTM and global max-pooling layers are then forwarded to a small feed-forward network for final prediction. This feedforward network consists of a fully-connected (dense) layer, followed by dropout and batch-normalization layers to avoid over-fitting~\cite{srivastava2014dropout, ioffe2015batch}. We use \textit{``ReLU"} activation function for CNN as well as fully-connected layers, which is defined by $\sigma(x) = max(x,0)$. Finally, a \textit{``softmax"} layer is used to get probabilities for each class and a label is assigned based on maximum probability. Figure~\ref{fig:architecture} illustrates the complete model architecture.

\section{Experimental Evaluation} \label{sec:evaluation}

\subsection{Dataset Description} \label{subsec:datasetDescription}
In this study, we use two publicly available datasets, having English and German languages respectively. The Sanders Twitter corpus~\footnote{https://github.com/zfz/twitter\_corpus} (hereinafter called as Twitter dataset) is in English language and has $5,513$ tweets, manually annotated as \textit{positive, neutral, negative} or \textit{irrelevant} classes. Note that there is no standard train/test split for this particular dataset. Hence, we follow the strategy employed in~\cite{Attia_2018_b9} and randomly split the $20\%$ of the data for testing purposes. We employ stratify sampling method to maintain the ratio of classes in the test split.

The second dataset is in German language for the task of aspect-based sentiment detection~\footnote{https://sites.google.com/view/germeval2017-absa/data} (hereinafter referred to as GermEval dataset). The standard training, development, testing 1, and testing 2 splits are available. Despite two test splits, only first split is used for testing in the literature, thus we also opt to use first split for testing. This dataset has only three classes i.e. \textit{positive, neutral} and \textit{negative}.

We also combine these two datasets to form a \textit{Mixed language} dataset for the evaluation of language independence of the proposed model. Note that we do not include the tweets marked as \textit{irrelevant} from the Twitter dataset in this combined dataset to maintain consistency across both languages. Class distribution with respect to each split of all datasets is given in Table\ref{tab:classDistributions}. We perform all the experiments without any preprocessing or over-sampling.

\begin{table}[!htbp]
	\centering
	\setlength{\tabcolsep}{3.5pt}
	\renewcommand{\arraystretch}{1.1}
	\caption{Class distribution of the datasets}\label{tab:classDistributions}
	\begin{tabular}{ccccccc}
		
		\toprule
		\textbf{Dataset} &  \textbf{Split} & \textbf{positive} & \textbf{neutral} & \textbf{negative} & \textbf{irrelevant} & \textbf{Total}\\
		\toprule
		
		
		Twitter  &  Train & $415$ & $1,866$ & $458$ & $1,351$ & $4,090$ \\
		&  Test & $104$ & $467$ & $114$ & $338$  & $1,023$\\
		
		\midrule
		GermEval & Train & $1,246$ & $14,497$ & $5,228$ & $-$   & $20,941$ \\
		& Dev & $149$ & $1,637$ & $589$ & $-$  & $2,375$ \\
		& Test-1 & $105$ & $1,681$ & $780$ & $-$   & $2,566$ \\
		& Test-2 & $108$ & $1,237$ & $497$ & $-$   & $1,842$ \\
		
		\midrule
		Mixed & Train & $1,631$ & $16,363$ & $5,686$ & $-$   & $23,680$ \\
		& Test & $209$ & $2,148$ & $894$ & $-$   & $3,251$ \\
		
		\bottomrule
	\end{tabular}
\end{table}

\subsection{Hyperparameters} \label{subsec:hyperparameters}
We empirically decide the number of CNN filters, kernel size $(k)$ of the CNN filters, LSTM units, and units in Dense layers. The value of $k$ in the CNN layer is set to $7$ to learn $7$-gram features. The rest of the hyperparameters are selected using a grid-search on the test split of the Twitter dataset. The selected hyperparameters remain the same for all other experiments. The hyperparameters, available choices, and selected values are presented in Table~\ref{tab:hyperparameters}.

\begin{table}[!tbp]
	\centering
	\setlength{\tabcolsep}{3.5pt}
	\renewcommand{\arraystretch}{1.1}
	\caption{Hyperparameters, available choices, and optimal value selected by grid search}\label{tab:hyperparameters}
	\begin{tabular}{ccc}
		
		\toprule
		\textbf{Hyperparameter} &  \textbf{Value Choices} & \textbf{Optimal Value}\\
		\toprule
		
		
		Dropout rate  &  0.1, 0.2, 0.3, 0.4, 0.5 & 0.5 \\
		Optimizer &  Adadelta, Rmsprop, Adam & Rmsprop  \\
		Learning rate &  0.001, 0.002, 0.003, 0.004 & 0.001 \\
		
		\bottomrule
	\end{tabular}
\end{table}
\subsection{Evaluation Metrics} \label{subsec:evaluationMetrics}
As seen from the Table~\ref{tab:classDistributions}, for each dataset, the distribution of number of instances in each class is very skewed, where \textit{``neutral"} class is dominant by a considerable margin. In such cases, \textit{Accuracy} is not an optimal measure of performance. Therefore, we choose to evaluate the results in terms of macro F1-score. However, for completeness sake, we also report \textit{Accuracy, Precision}, and  \textit{Recall} for all variations of the experiments.

\subsection{Results and Discussion} \label{subsec:results}

We focus the discussion on the results with respect to F1-score. However, other metrics are also reported in Table~\ref{tab:results}. We take the language-independent model presented in~\cite{Attia_2018_b9} as baseline. For mixed languages, we re-run the baseline model as this specific setting of mixed language was not reported in the original study. We perform two kinds of experiments with respect to the choice of embedding dimensions. First, we choose $d = 300$, as suggested in~\cite{Attia_2018_b9}. Then we also perform the experiments by reducing the embedding dimensions $d$ to $100$ and report the performance of the proposed model.

%
%
%
%
%
%
%
\begin{figure*}[!t]
	\begin{tabular}{ccc}
		\centering
		\includegraphics[width=.318\linewidth]{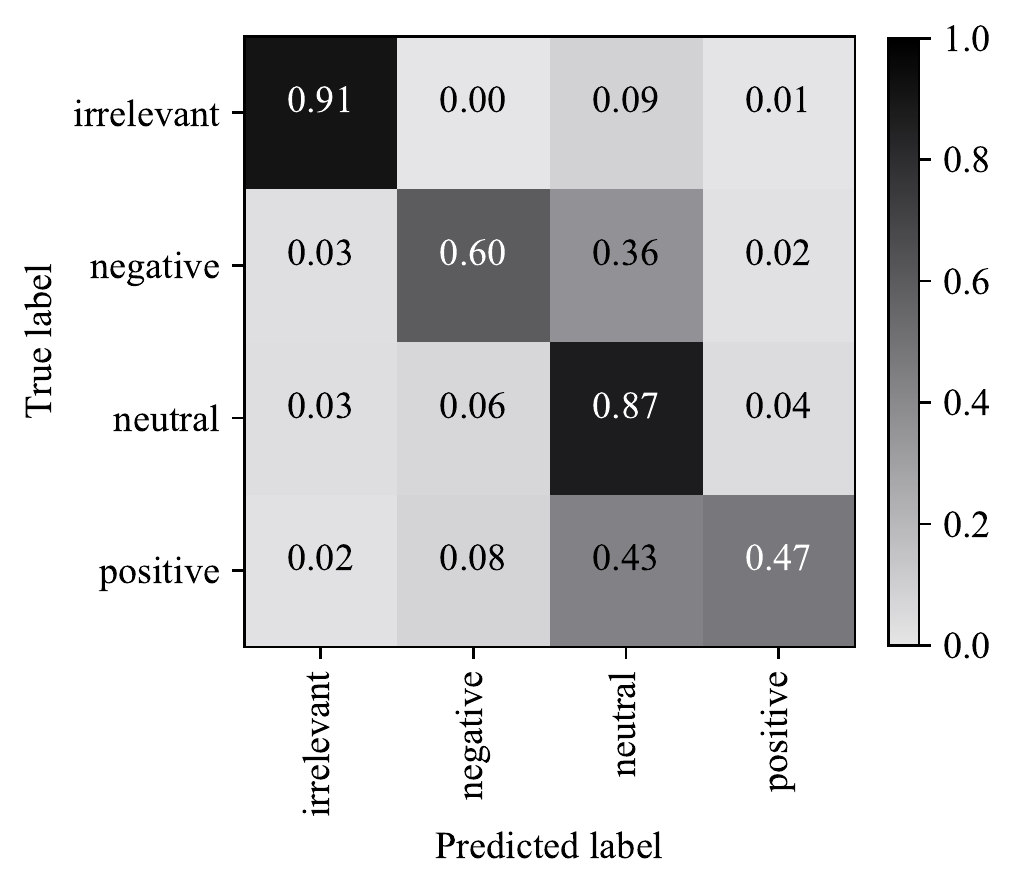}&
		\includegraphics[width=.318\linewidth]{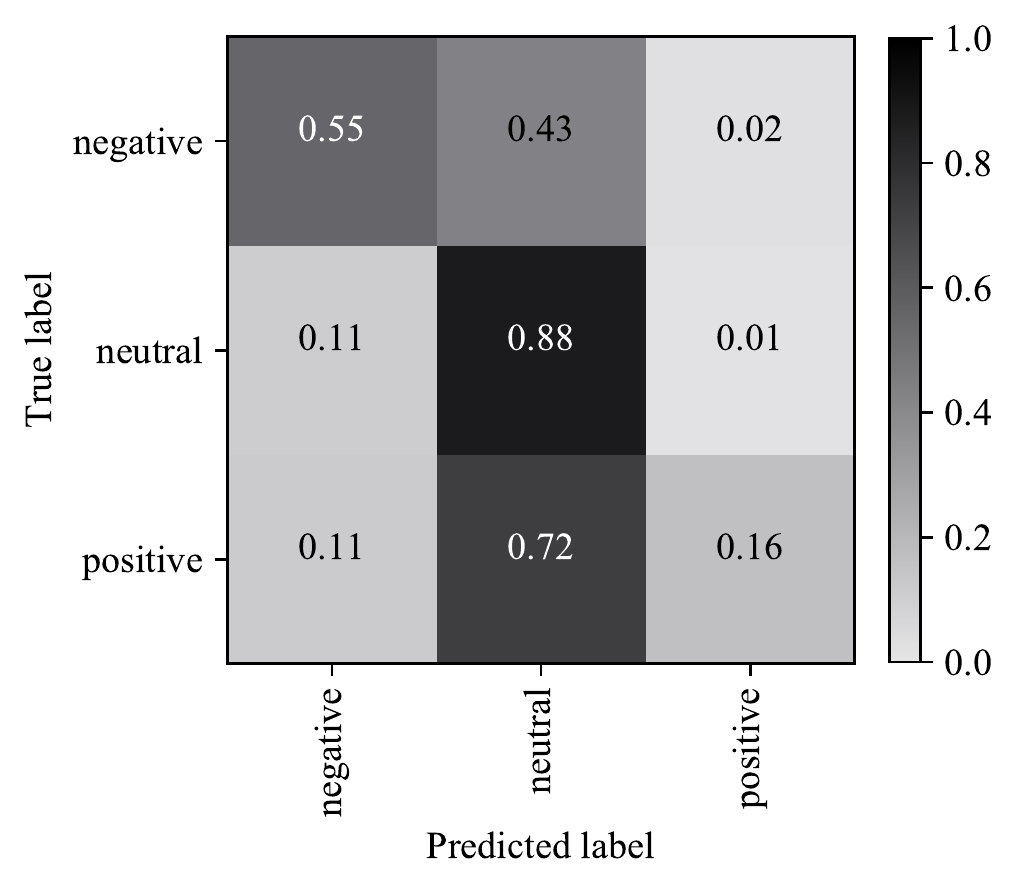}&
		\includegraphics[width=.318\linewidth]{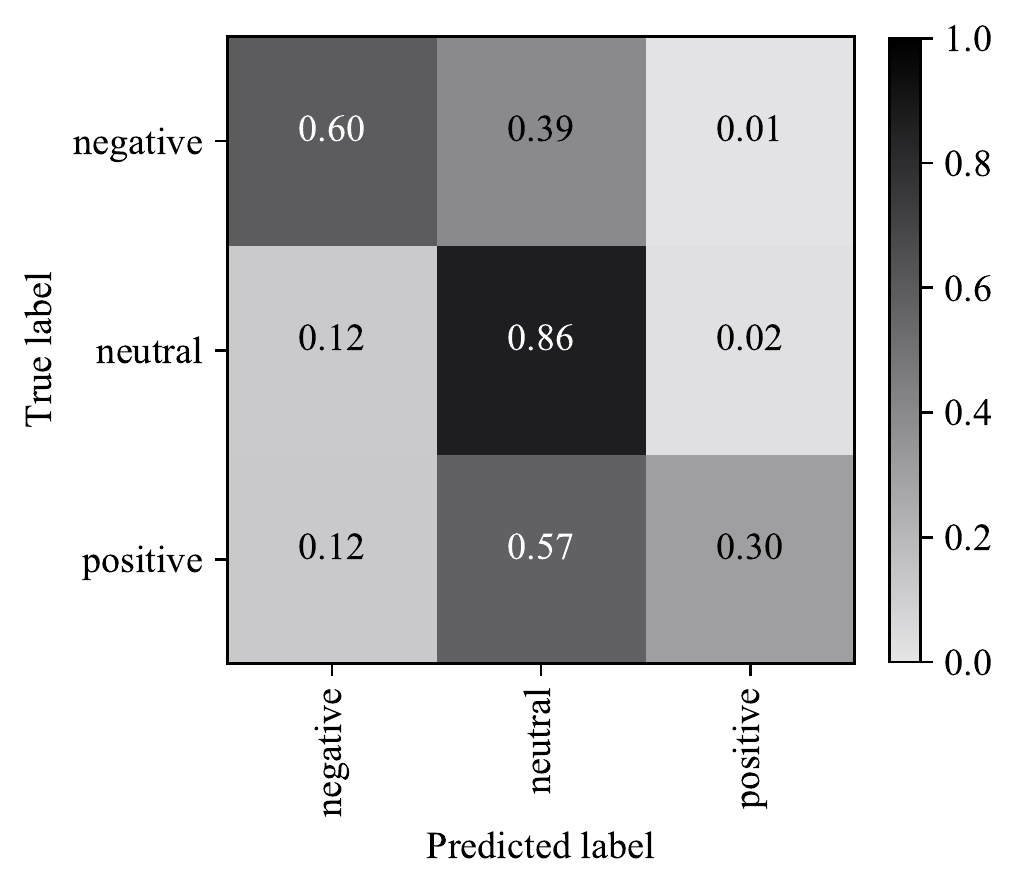}\\
		\hspace{1.4mm}  (a) English  & \hspace{1.4mm}  (b) German & \hspace{1.8mm} (c) Mixed \vspace{1.5mm} \\
		
	\end{tabular}
	\caption{Confusion matrix for prediction on test split of each language}
	\label{fig:ConfusionMatrix}
\end{figure*}

With respect to Twitter dataset, traditional Machine Learning algorithms, such as SVM, Decision Trees, Naive Bayes, and Logistic Regression were applied by~\cite{bravo2013combining}. They reported that they achieved $70.10\%$ testing accuracy. Another approach presented in~\cite{hassan2013twitter} used an ensemble of multiple learning algorithms and utilized majority vote to get the final prediction. This approach was able to achieve an accuracy of $76.30\%$. A similar approach was opted in~\cite{da2014tweet} which achieved comparable accuracy of $76.25\%$. More recently, the baseline model based on CNN was able to achieve an accuracy of $78.60\%$ with F1-score of $69.13\%$. They also perform the experiments by oversampling the underrepresented classes and report an accuracy of $79.57\%$ while F1-score with oversampling was increased to $70.23\%$. It is worthwhile to note that the proposed model outperforms the baseline for both embedding dimensions choices. However, highest F1-score is achieved with $300$ dimensions. With respect to F1-score, our model shows a significant improvement of $2.77\%$ over the baseline. Note that the results indicate that the proposed model has an improved ability to differentiate between the classes without any preprocessing and oversampling. This proves that the model is robust with respect to skewness in class label distribution, as evident from Figure~\ref{fig:ConfusionMatrix} (a).

\begin{table}[!tbp]
	\centering
	\setlength{\tabcolsep}{4.5pt}
	\renewcommand{\arraystretch}{1.5}
	\caption{Performance evaluation of variations of the proposed model and baseline. Showing highest scores in boldface.(subscript ``O" stands for oversampling)}\label{tab:results}
	\begin{tabular}{llcccc}
		
		\toprule
		\textbf{Language} & \textbf{Model} & \textbf{Accuracy} & \textbf{Precision} & \textbf{Recall} & \textbf{F1-score} \\
		
		\toprule
		Twitter & Baseline~\cite{Attia_2018_b9} & $78.60$ & $-$ & $-$ & $69.13$ \\
		& Baseline$\textsubscript{O}$~\cite{Attia_2018_b9} & $79.57$ & $-$ & $-$ & $70.23$ \\
		& Proposed$_{300}$ &  $\textbf{80.94}$ & $\textbf{76.14}$ & $\textbf{71.00}$ & $\textbf{73.00}$ \\
		& Proposed$_{100}$  &  $79.57$ & $75.64$ & $69.22$ & $71.70$ \\
		
		\midrule
		GermEval & Baseline~\cite{Attia_2018_b9} &  $\textbf{75.45}$ & $-$ & $-$ & $49.00$ \\
		& Baseline$\textsubscript{O}$~\cite{Attia_2018_b9} &  $73.73$ & $-$ & $-$ & $\textbf{54.84}$ \\
		& Proposed$_{300}$ &  $74.75$ & $\textbf{58.57}$ & $\textbf{52.90}$ & $54.79$ \\
		& Proposed$_{100}$  &  $74.79$ & $61.17$ & $49.30$ & $51.27$ \\
		
		\midrule
		Mixed & Baseline~\cite{Attia_2018_b9} &  $74.75$ & $-$ & $-$ & $57.37$ \\
		& Proposed$_{300}$ &  $\textbf{75.08}$ & $\textbf{66.50}$ & $\textbf{58.53}$ & $\textbf{61.24}$ \\
		& Proposed$_{100}$  &  $74.93$ & $64.68$ & $58.13$ & $60.63$ \\
		
		\bottomrule
	\end{tabular}
\end{table}

In regards to GermEval dataset, we report results with respect to test split $1$. On this test split, the results of multiple algorithms are reported in~\cite{wojatzki2017germeval}. SWN2-RNN approach achieved $74.9\%$ accuracy while fasttext based embedding were able to achieve $74.8\%$ accuracy. A hybrid approach based on SVM and biDirectionalLSTM was able to achieve an accuracy of $74.5\%$. The baseline model without any oversampling reports the best results in literature on this specific split of GermEval dataset, which is $75.45$. However, the F1-score without oversampling is $49.00\%$. The proposed model is able to achieve an F1-score of $54.79\%$, which is significantly higher than baseline model (without oversampling). When the strategy of oversampling is incorporated in~\cite{Attia_2018_b9}, the F1-score attained a significant boost on the expense of accuracy. Our model achieves comparable results with oversampled baseline model.

Turning now to the case of mixed language, no results in literature exist on this specific setting. However, we re-run baseline model on mixed language dataset. The model was able to achieve an accuracy of $74.75\%$ with F1-score of $57.37\%$. However, the proposed model with $300$ dimensional embedding vectors significantly outperformed the base model with F1-score of $61.24\%$. The results lead us to conclude that the proposed model exhibits language independence.

Figure~\ref{fig:ConfusionMatrix} shows confusion matrix for test split of all three datasets using the proposed model (with $300$ dimensional embeddings). The figure shows that the proposed model favors \textit{``neutral"} class, which makes intuitive sense as the number of instances in this class are significantly higher than other classes.

\section{Conclusions} \label{sec:conclusion}

In this study, we presented a novel architecture for the problem of language independent sentiment analysis. We use two datasets of English and German languages, while we also perform experiments by combining both datasets to evaluate the language independence of the model. We compare our results with existing state-of-the-art model on these datasets and show that the proposed model outperforms the baseline by a significant margin with respect to F1-score on English and mixed language datasets. Our model achieves comparable results on German language dataset. The results show that the proposed model can successfully be applied to multiple or mixed languages, without major loss in predictive performance.

\bibliographystyle{IEEEtran}
\bibliography{sentAnalysi_Bib}

\end{document}